\documentclass[journal]{IEEEtran}
\usepackage{graphicx}
\usepackage[numbers,sort&compress]{natbib}

\ifCLASSINFOpdf
\else
\fi
\begin{document}
\title{QwenGrasp: A Usage of Large Vision-Language Model for Target-Oriented Grasping}  

\author{Xinyu Chen, Jian Yang*, Zonghan He, Haobin Yang, Qi Zhao, Yuhui Shi*% <-this % stops a space
\thanks{The authors were with the Department of Computer Science and Engineering, Southern University of Science and Technology, Shenzhen, China, 518055.}% <-this % stops a space
\thanks{Correspondence: Jian Yang, Yuhui Shi.}
\thanks{Email: yangj33, shiyh@sustech.edu.cn.}
}
\maketitle


\begin{abstract}
Target-oriented grasping in unstructured scenes with language control is essential for intelligent robot arm grasping. The ability of the manipulators to understand the human language and execute corresponding grasping actions is a pivotal challenge. 
This paper proposes a combination model called QwenGrasp, which combines a large vision-language model with a 6-DoF grasp neural network and can conduct a 6-DoF grasping task on the target object with textual language instruction.
We conducted several experiments with six-dimension instructions to test the QwenGrasp when facing different cases. The results show that QwenGrasp has a superior ability to comprehend human intentions. Even in the face of vague instructions with descriptive words or instructions with direction information, the target object can be grasped accurately. When QwenGrasp accepts an instruction that is not feasible or irrelevant to the grasping task, our approach can suspend the task execution and provide proper feedback to humans, improving safety.
In conclusion, with the great power of the large vision-language model, QwenGrasp can be applied in the open language environment to conduct the target-oriented grasping task with freely input instructions. 
Our videos are shown at \emph{https://chuxiyeok.github.io/IntelligentGrasping/}
%\emph{https://github.com/ChuxiyeOK/IntelligentGrasping.}
\end{abstract}

%\begin{IEEEkeywords}
%IEEE, IEEEtran, journal, \LaTeX, paper, template.
%\end{IEEEkeywords}

% \fi
\IEEEpeerreviewmaketitle

\section{Introduction}
%\IEEEPARstart{}{}

Target-oriented grasping enables a manipulator to grasp target objects under human intention, thereby enhancing the efficiency and safety of grasping tasks \cite{do_as_i_can}.
In the open world of coexisting humans and robots, it is expected that the robot arm could conduct the target-oriented grasping task by inputting textual instructions from humans. Humans make mistakes, and robots should be able to determine the feasibility of a given task autonomously. It will avoid executing the task mindlessly with erroneous instructions.
The target-oriented grasping task encompasses two primary abilities: the capability to understand human instructions \cite{joint} and the ability to ensure a high success rate in grasping \cite{regnet}. Both of these aspects are crucial for the successful execution of the target-oriented grasping task.

Some previous works achieved the target-oriented grasping task by matching human intentions with the target object according to a given template image \cite{t1,t2,t3,t4}. However, these approaches are not user-friendly and often demand high-quality image information, which lacks generality, making it impractical in many situations without target images. Recent research efforts in target-oriented grasping with textual instructions aim to bridge the gap between the human language and the process of grasping target objects \cite{do_as_i_can, joint,seeask}. For example, some studies have harnessed the capabilities of the visual-language model Contrastive Language Image Pre-training (CLIP) \cite{clip} to facilitate the matching of text and images, thereby establishing a mapping from human instructions in text to the corresponding target objects for robot arm manipulation. 

However, it has to be noted that the CLIP cannot comprehensively understand the scene with spatial context, which introduces the limitation. Moreover, the CLIP lacks the flexibility to handle complex human instructions in a real-world environment. These CLIP-based methods may execute grasping tasks mindlessly when confronted with infeasible or non-grasping instructions. The grasp system of these methods is limited to the specific instruction templates. They are prone to unexpected problems when applied in the daily natural language environment with many different instructions.

Ensuring a high success rate in grasping is also a crucial aspect of target-oriented grasping. Previous works \cite{regnet,grasp1b,mono} have achieved a high successful rate in 6-DoF grasping in unstructured scenes. They \cite{s4g,grasp1b,6dofgrasp} often extract point cloud data from the scene and employ the PointNet \cite{pointnet} as a backbone neural network to generate candidate grasp poses for objects in the scene. Furthermore, they \cite{regnet, pointnet} also use the PointNet \cite{pointnet} to evaluate the generated grasp poses then pick out the best pose. However, these methods primarily focus on the grasping success rate and overlook the target selection. For these methods, the final grasping target may exist on any object in the scene, which means their grasping target is unknown and they can not be directly used in target-oriented grasping task.

Recent researches \cite{palm-e,voxposer} have been increased using multi-modal large language models, specifically the Large Vision-Language Models (LVLMs) to control robots through simple language instructions. They have demonstrated a remarkable understanding of human intention and achieved impressive results. These works \cite{palm-e,voxposer} have bestowed robots with a level of comprehension comparable to that of humans. And these LVLMs assist the robot to easily understand human intention and autonomously control robotic systems to accomplish corresponding tasks, resulting in astonishing outcomes. However, while these achievements have proven successful in relatively straightforward grasping tasks, their specific effectiveness in more complex scenes such as target-oriented 6-DoF grasping in unstructured scenes, remains to be rigorously tested and evaluated.

\begin{figure}[t]
\includegraphics[scale=0.32]{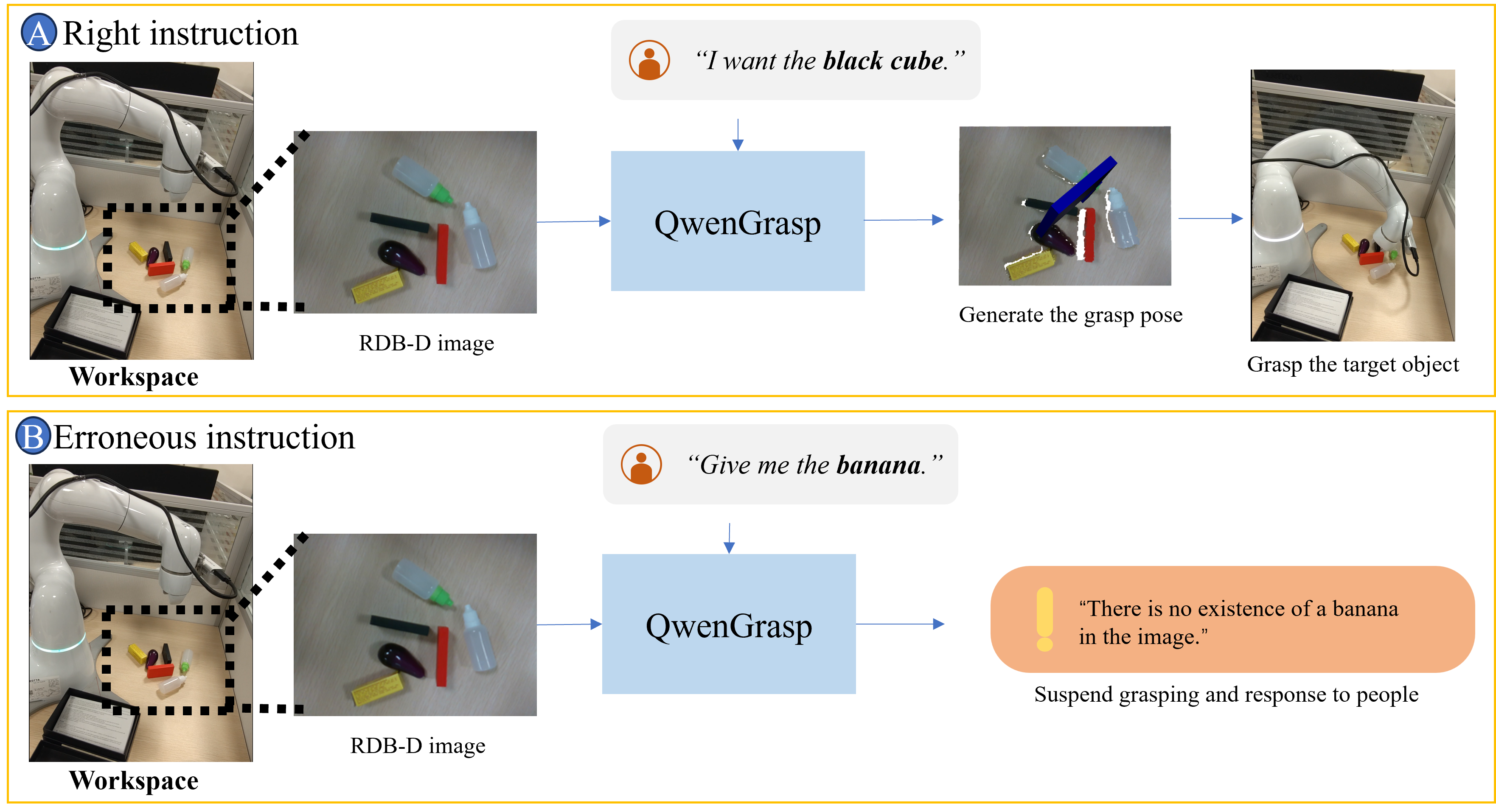}
\caption{Two common cases of target-oriented grasping with right or erroneous instruction. In the case A with right instruction at the top half of this picture, our combination model QwenGrasp will conduct grasping to the target object as the willing of human. In the case B with erroneous instruction at the bottom half of this picture, QwenGrasp will realize the infeasible task and suspend the grasp mission.}
\vspace{-1em}
\label{fig.1}
\end{figure}  

In this paper, we propose a novel combination model called QwenGrasp for target-oriented grasping tasks. QwenGrasp is composed of the large vision-language model Qwen-VL \cite{qwenvl} and the grasping neural network REGNet \cite{regnet}. Specifically, we use the pre-trained large vision-language model Qwen-VL \cite{qwenvl} to encode the RGB image of the workspace and the textual human instruction. And it can output a detection bounding box on input image for the detected target object. REGNet \cite{regnet} is a grasping neural network specialized for 6-DoF grasping in unstructured scene. We then employ the pre-trained REGNet to generate the final grasp pose based on the target object. Since we directly use the pre-trained large vision-language model and pre-trained grasping neural network, our combination model QwenGrasp requires no additional training.

Compared to previous works, not only QwenGrasp can complete the target-oriented grasping task by textual instructions, but also it has the ability to determine the feasibility of the given instruction. QwenGrasp demonstrates the effectiveness in two key abilities of target-oriented grasping tasks: understanding textual human language instructions and ensuring the grasp success rate. Particularly, Qwen-VL \cite{qwenvl} is one of the leading large vision-language models, it offers the best ability for understanding human textual instructions and workspace spatial information. As a result, our proposed method can handle complex and varied language instructions, achieving language generalization close to everyday human conversations. Besides, as shown in Fig. \ref{fig.1} case B, when it facing with erroneous instruction, our method can suspend tasks and ask human for confirmation, greatly enhancing the safety of robot arm grasping.

To summarize, our main contributions are as follows:
\begin{itemize}
\item[$\bullet$] We propose a combination model called QwenGrasp that combines a large vision-language model with a professional grasp network for the target-oriented 6-Dof grasping in unstructured scenes. To the best of our knowledge, we are the first to use the multi-modal large language model in target-oriented 6-DoF grasping with textual instructions. Also the first to use Qwen-VL in grasping.
\item[$\bullet$]We modify the REGNet to fit Qwen-VL, combining the bounding box of target object. Also, We design a comprehensive set of prompts to preload in the large vision-language model, strengthen the robot arm with the ability to autonomously assess task feasibility, and independently plan and execute grasping tasks.
\item[$\bullet$] Our proposed methodology is evaluated across a range of real-world scenes encompassing common objects coupled with natural language instructions. The outcomes validate its effectiveness and generalizability.
\end{itemize}

The remainder of the paper is arranged as follows. Section 2 summarizes the related works of target-oriented grasping, 6-DoF Grasping in unstructured scenes and multi-modal large language models for robotic grasping. Section 3 introduces our methodology and implementation details. Section 4 shows the experiment results of our method with six kinds of instructions. Finally, a conclusion including future works is placed in Section 5.

% I wish you the best of success.

 

\section{Related Work}

\begin{figure*}[t]
\centering
\includegraphics[scale=0.5]{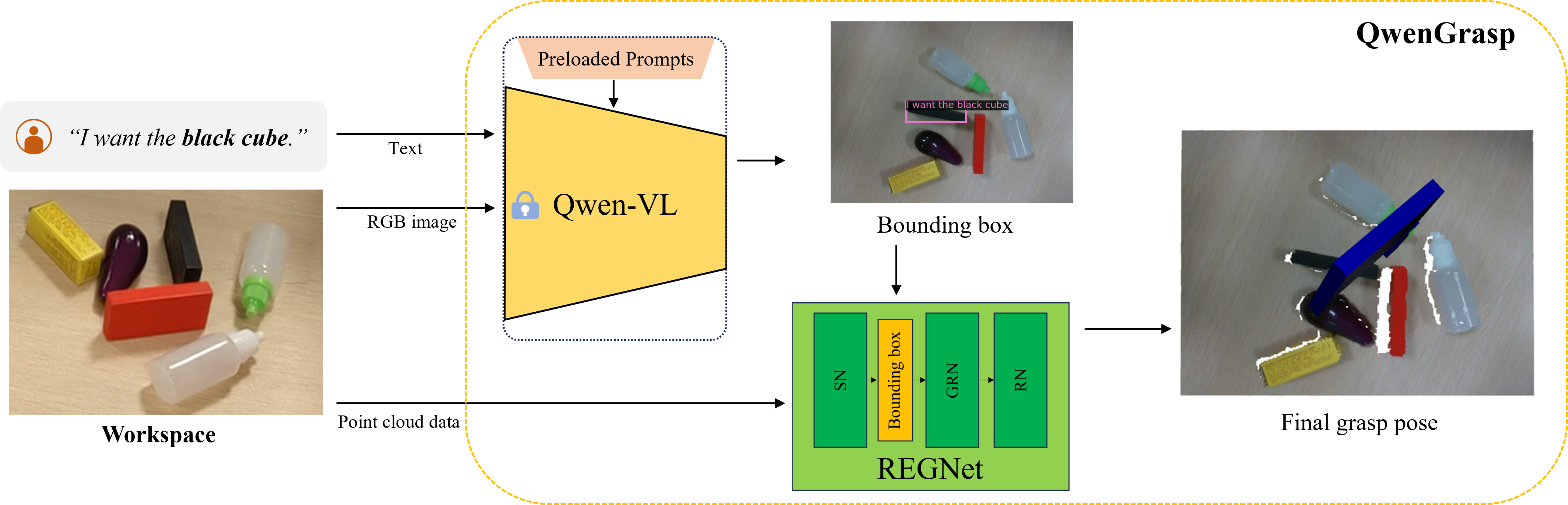}
\caption{ \textbf{QwenGrasp Overview}. Given the textual instruction, our system will acquire the RBG image and the point cloud data of the workspace. The pre-trained Qwen-VL \cite{qwenvl} will generate the bounding box of the target object, and the REGNet \cite{regnet} will generate the candidate grasp poses of the scene. To generate the final grasp pose in the desired place, the REGNet is modified to combine the bounding box parameters from Qwen-VL.}
\vspace{-1em}
\label{fig.2}
\end{figure*}

\subsection{Target-oriented grasping}
The robot arms target-oriented grasping with human textual instructions has always been a significant task, offering substantial benefits in both industrial production and daily life \cite{joint}. Recent years have witnessed numerous explorations in this area \cite{t1,t2,t3,t4}. Recent works \cite{joint,seeask} have leveraged CLIP \cite{clip} to achieve the target-oriented grasping by textual language. Vision-Language-Action \cite{joint} can grasp blocked targets by grasping away the obstacles. SeeAsk \cite{seeask} can help the robot identify a target object by answering its questions. They all use preset language templates for training or questioning. CLIP \cite{clip} is a multi-modal large-scale visual-language model to matching the text and image. By directly using pre-trained CLIP, these works achieve matching texts and objects without additional training. By combining the CLIP with a grasping network, it realize an efficient process for translating human instructions into target-oriented grasping, yielding favorable results in specialized grasping environments. However, CLIP lacks the ability to comprehend spatial relationships among objects in the workspace and falls short of the language understanding capabilities of large language models (LLMs) like the ChatGPT \cite{chatgpt3} and so on \cite{llama2}. Consequently, CLIP-based target-oriented grasping methods may not deliver satisfactory results when dealing with complex and flexible instructions, and these instructions can be frequently encountered in real-world applications.

\subsection{6-DoF Grasping in unstructured scenes}
The exising methods \cite{mono,regnet,pointnetgpd} of 6-DoF Grasping in unstructured scenes reaches an outstanding enough grasping success rate. 6-DoF grasping makes the robot arm grasp objects on the workspace from any height and direction. The robot arm gripper pose contains the 3D gripper position and the 3D gripper orientation. Compared with 2D planar grasping, 6-DOF grasping has a higher grasping success rate in unstructured scenes with unseen objects. Except for the work \cite{mono} using the RGB image as input, most works \cite{s4g,grasp1b,6dofgrasp} take the 3D point cloud data of the workspace as input. They will generate candidate grasp poses based on the point cloud and analyse the grasp stability and success rate by giving a score. Finally, the grasp pose with the highest score can be the grasp pose for the robot arm to execute grasping. The REGNet \cite{regnet} has three stages: Score Network (SN), Grasp Region Network(GRN), and Refine Network (RN). SN selects the 3D gripper positions. GRN generates the 3D gripper orientations. RN refines those grasp poses and gives the final result. The GraspNet-1Billion \cite{grasp1b} makes a grasp dataset with 1 billion labeled grasp poses including a wide range of general objects. These 6-DoF grasping methods can be used to grasp novel objects and conducted in unstructured scenes.
%\IEEEpubidadjcol
\subsection{Multi-modal Large Language Model}
The development of large models has greatly benefited people's work and daily life. Recent works \cite{palm-e,voxposer} have combined these large models with robots, enabling robots to carry out various tasks in response to human commands.

PaLM-E \cite{palm-e} is a large model which is composed of vision transformer and the large language model PaLM \cite{palm}. It's an embodied multi-modal language model. It accepts multi-modal inputs including images, 3D models, and text, automatically analyzes the given queries and generates subtask sequences. While maintaining the effective communication capabilities of large language models, it enables robots to perform a wide range of tasks guided by natural language instructions. VoxPoser \cite{voxposer} is also based on large language model. It has the abilities to synthesize robot trajectories with the intention of natural language instructions. And it can directly generate the corresponding code to control robots for a diverse set of manipulation tasks. However, these works have not been specifically applied for grasping tasks.

In this paper, we use the large vision-language model Qwen-VL \cite{qwenvl} to help robots understand human instructions and locate target objects. Qwen-VL \cite{qwenvl} is a multi-modal large language model based on the large language model QwenLM \cite{qwenvl}. It can accept both images and text as input. Besides the ability of image captioning, question answering, and flexible interaction, it has a strong visual localization ability, which can locate objects in images by understanding the textual instructions. The main part of our approach QwenGrasp is to use Qwen-VL to understand human instructions and generate the detection bounding box to capture target objects. This overcomes one of the most important obstacles of the target-oriented grasping problem, which is freely understanding human instructions. By combining Qwen-VL, our method can be applied in people's daily-life language environment and help people realize the grasping of target objects. Even in the face of instructions not to grasp or wrong, it can realize the human mistake and autonomously reply an intelligent response, rather than blindly performing the grasp task.

\section{Method}

\begin{figure*}[t]
\centering
\includegraphics[scale=0.6]{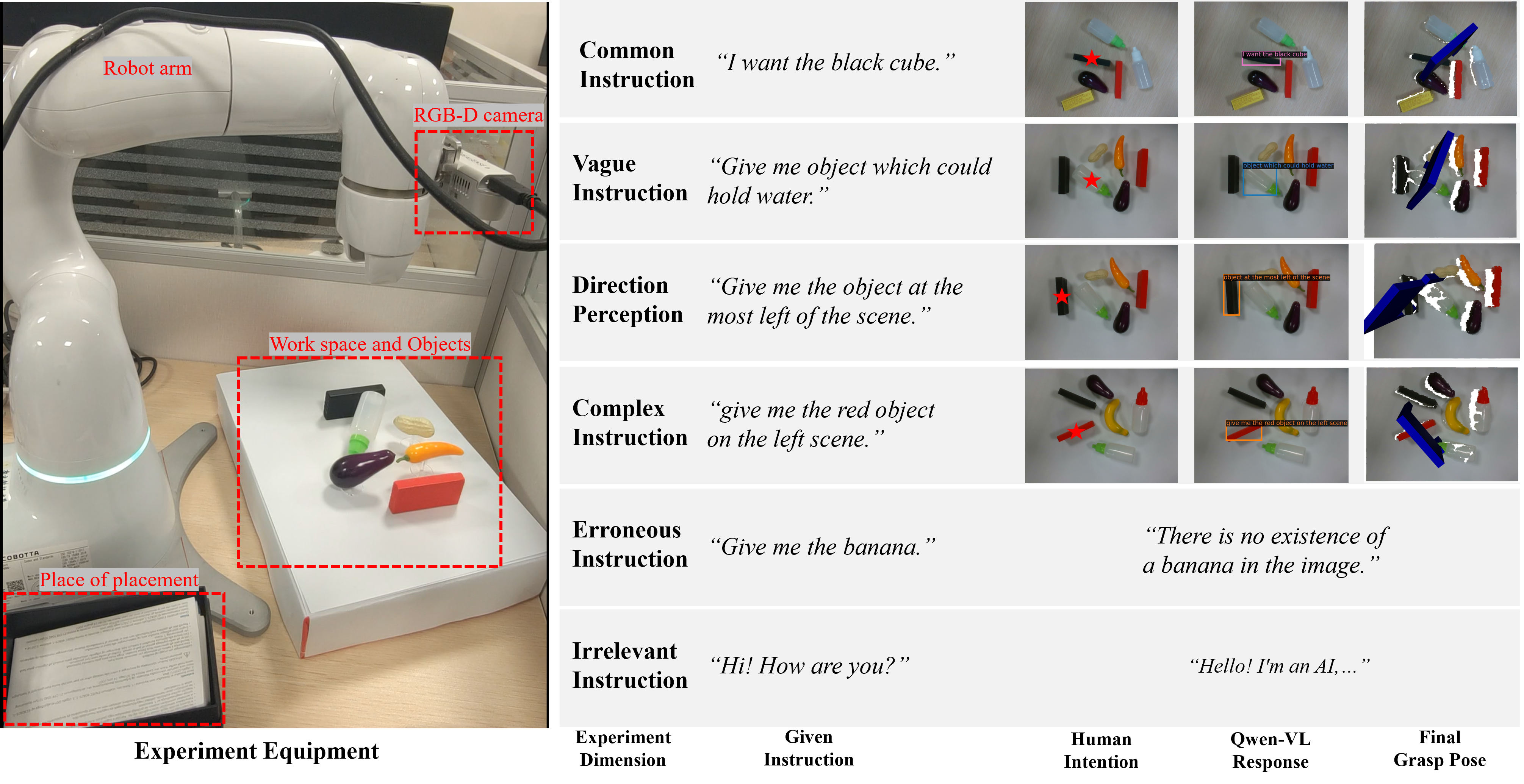}
\caption{ \textbf{Six-Dimension Experiments}. On the left side of the whole picture, it shows the experiment equipment and the workspace. We use COBOTTA robot arm with an Intel RealSense D435 RGB-D camera. Some common seen objects in daily life are placed on the work space. On the right side of the picture, it shows the examples of 6-dimension experiments.  Each of the experiments shows a different kind of instruction. We use a red five-pointed star to flag the human attention. Show the response from Qwen-VL by drawing the bounding box on the image. The final grasp pose selected is shown on the point cloud data with a blue grasp pose. For the special cases of Erroneous Instruction and Irrelevant Instruction which there is no target object on the image, we show the textual response of Qwewn-VL.}
\vspace{-1em}
\label{fig.3}
\end{figure*}

\subsection{QwenGrasp Overview}
As shown in Fig. \ref{fig.2},  human give an instruction to the QwenGrasp system to order the object he want. The system automatically calls a RGB-D camera to get the RGB image and point cloud data of the workspace. After confirming the correctness and feasibility of the instruction, the QwenGrasp system will output the final grasp pose of the target object. Our proposed QwenGrasp is mainly composed of Qwen-VL \cite{qwenvl} and REGNet \cite{regnet}. Qwen-VL accepts human instructions and RGB images of the workspace as inputs. It can understand textual human language and images, match objects in the image with human instructions, and output a bounding box based on the input RBG image to represent the location of the target object. Preloaded prompts is the prompts input in Qwen-VL before we use the QwenGrasp, aiming to make Qwen-VL suit for the grasping task. REGNet \cite{regnet} is a 6DoF grasping network, which is modified here to combine the Qwen-VL output. By inputting the 3D point cloud data of the scene to the REGNet, candidate grasp poses can be generated based on the workspace. Finally, the modified REGNet will utilize the detected bounding box to filter the target point cloud data to generate the final grasp pose.

\subsection{Preloaded Prompts }
Since Qwen-VL \cite{qwenvl} is a versatile vision-language model, in order to make it more suitable for target-oriented grasping tasks, we input the prompts designed in advance to the model. Our preloaded prompts have two main purposes: 1) Guide the versatile large model to match the given instruction and images, and locate the target object; 2) Limit the output of the large model to make it focus on grasping task. We don't set any restrictions on human instructions, and allow the large model to flexibly respond to a wide variety of human inputs. With preloaded prompts, the model classifies human instructions into three categories: 1) This instruction is a target-oriented object grasp instruction, and QwenGrasp can find the target; 2) This instruction is an erroneous target-oriented object grasp instruction, and there is no target object in the workspace; 3) This instruction is an irrelevant instruction to the target-oriented grasping task. With the guidance of preloaded prompts, Qwen-VL will directly output the detected bounding box of the target object when accepting the correct task instruction. In the face of erroneous instructions or irrelevant instructions to the grasping task, the detection bounding box will not output. On contrast, the human will be reminded by QwenGrasp to check whether the input is correct.

\subsection{Pre-trained Models}
Qwen-VL \cite{qwenvl} is a large vision-language model and the pre-trained Qwen-VL is used in our method. We get the bounding box of the target object by Qwen-VL. Qwen-VL is based on the large language model Qwen-7B. Qwen-VL has a total of 9.6B parameters, in which is the sum of the parameters in Qwen-7B, the vision transformer and cross attention model. Qwen-VL accept text and images as input. It has a remarkable performance in image captioning, question answering, visual localization, and flexible interaction tasks. Our QwenGrasp takes the advantage of Qwen-VL's powerful visual localization capabilities to locate the place of the target object. By matching the instruction and specified grasping target on the input RGB image of the workspace, QwenGrasp obtains the 2D spatial position of the target object. It solves one of the most serious problems that is locating the position of the target object on the image with textual instruction in target-oriented grasping. Meanwhile, as a large vision-language model, Qwen-VL has a strong dialogue ability and image understanding ability. It can also respond well to complex or flexible instructions. It can accept any human instruction and respond to it. Instead of blindly executing tasks due to erroneous instructions, QwenGrasp will remind humans of the error and give suggestions.

REGNet \cite{regnet} is a 6-DoF grasping network and we use the pre-trained REGNet to guide robot arm grasping. It could be used in unstructured scenes and deal with novel objects. REGNet \cite{regnet} has a 3-level structure, including the Score Network (SN), Grasp Region Network (GRN), and Refine Network (RN). Such structure ensures the diversity and stability of candidate grasping poses, and ensures the grasping success rate. Our QwenGrasp utilizes REGNet to generate a large number of candidate grasp poses. Each candidate grasp pose has a corresponding score representing the grasp quality, which greatly facilitates the final grasp pose selection.

\subsection{The Modified REGNet}
the REGNet is originally used to generate the 6-DoF grasp poses, and we modified it to to generate the proper 6-DoF grasp pose of the target object. REGNet has a 3-level structure, which is composed of the SN, GRN and RN. The bounding box feature of the target object from Qwen-VL is added between the SN and GRN. SN select the best group of points from input point cloud data, then the bounding box information is used to further filter the points belonging to the detected range. After that, the remained points are send to the next stage of the REGNet, and the grasp poses of the target object are generated. Finally, according to the scores of the remaining candidates grasp poses, the final grasp pose is selected. The grasp pose with the maximum score will be selected as the final grasp pose.

\section{Experiments}

We evaluate QwenGrasp in the real world. As shown in Fig. \ref{fig.3}, our real world experiment platform involves a COBOTTA robot arm with a 2-finger parallel-jaw gripper to grasp objects and an Intel RealSense D435 RGB-D camera fixed on robot arm to acquire workspace RGB-D images of resolution 640 x 480. We prepared different objects that are common seen in house and office scenarios to test how our approach performs in a familiar environment for most humans. In order to better demonstrate the effect of our QwenGrasp, we designed a 6-dimension experiment to test it comprehensively. They are: 1) Common Instruction; 2) Vague Instruction; 3) Direction Perception; 4) Complex Instruction; 5) Erroneous Instruction; 6) Irrelevant Instruction. And we already have some preliminary results.

\subsection{Common Instruction}
The purpose of this experiment with common instruction is to test the basic functionality of QwenGrasp. In target-oriented tasks, our QwenGrasp needs to understand both the human instruction and the image of the workspace. It should complete the task of matching the human instruction text and the target object in image. In this experiment, we assume that humans have learned that the QwenGrasp is a target-oriented grasping combination model based on large vision-language model, so the input instructions are simple and straightforward. For example, "Give me the mug.", "please grasp me that black pen.". In these experiments, we demonstrate the ability to match images and instructions in a grasping environment. QwenGrasp has a remarkable object recognition ability and it's able to cope with common objects in daily life easily.

\subsection{Vague Instruction}
In many cases, we are not able to know the names of all the items. Even we may not be on-site in the workspace and we can only control the robot to grasp objects through a way of language control. In such cases, we can only describe our grasping target object through verbal descriptive words. For example, "Give me object which could hold water." "give me the longest object in that table." We call such instructions vague instructions. The purpose of this experiment is to test how would QwenGrasp choose in a grasping environment where the object was not explicitly specified. Our results show that QwenGrasp has a remarkable understanding ability and it always choose the most suitable object.

\subsection{Direction Perception}
Qwen-VL has the ability to understand images, so we have to test it for this feature. In some scenes, we may only know the names of some objects, and other objects are difficult to describe in words. In these cases, if the target object is next to the known object, then people can express their intention by couple the information of the known object with the location information. For example, "Give me the object between the mug and the bottle.", "give me the object at the lower right corner of the scene.". This experiment shows that QwenGrasp has a great direction perception ability in the grasping environment, and it can grasp the items just by combining direction information with object information.

\subsection{Complex Instruction}
The purpose of this experiment is to test QwenGrasp to understand the hidden human intention by input complex instructions. In the face of long sentences, instructions that contain misleading information, or complex instructions where there is only a small number of valid information hidden in a large number of irrelevant statements, we test the ability of the QwenGrasp whether it could recognize human intention and correctly find the target object to grasp. Complex instructions are typically long text. For example, "give me the red object on the left scene.", in this sentence, it contain both the vague description and direction perception information which greatly increased the difficulty for detecting the target object. In such experiments, The results demonstrate the abilities of QwenGrasp to understand complex sentences and extract human intentions in the grasping environment. It shows the ability to accurately match the target object with complex instruction.

\subsection{Erroneous Instruction}
Sometimes, human make the mistake of giving the target object which don't exist on the workspace. At that time, forcing the system to find a most similar item, blindly performing the grasp task will only lead to fatal errors. The correct approach should be suspending to grasp the object and explaining to the human that there is no target object in the workspace, please change the target. This experiment sets diverse items in the workspace, but the instruction will specify a different item. Giving erroneous instructions in an attempt to mislead the QwenGrasp. As expected, after conducting experiments, the results show that when faced with erroneous instructions, QwenGrasp can understand there is no target object that humans want in the scene. And it will not output the grasp bounding box of the target item, suspending the grasping task.

\subsection{Irrelevant Instruction}
Similar to the erroneous command, the robot arm should not blindly perform the grasping task when the human gives a command that is irrelevant to the grasping task. However, these instructions may be some QwenGrasp related questions. For example, "who are you? ", "what can you do?”. Faced with these questions, it’s expected to get an answer to help people use QwenGrasp. And there is another kind of irrelevant instructions, just caused by people wrongly send to the robot. Faced with these case, QwenGrasp should not execute grasp task but reply a proper response to people. The results of our experiments show that when faced with irrelevant instructions, QwenGrasp will suspend grasping tasks and communicate with humans according to the instructions to solve people’s questions.

\section{Conclusion}
In this work, we propose a combination model called QwenGrasp for target-oriented grasping in open-world. QwenGrasp shows remarkable performance in all six dimensions of the experiments. Not only it can understand human instructions, find target objects, and understand complex and vague instructions, but also it can perceive direction information of the workspace. Furthermore, in the face of erroneous instructions and irrelevant instructions, QwenGrasp can be aware of that and intelligently suspend the grasping task. Besides, QwenGrasp could communicate with humans, confirm the correctness of instructions and help humans solve problems, preventing blindly conduct the grasping task. The experiment results show that with the help of large vision-language model, QwenGrasp can easily detect and grasp objects on target-oriented grasping tasks. By combine the grasping task with the large vision-language model, people can freely use QwenGrasp, which significantly improves the safety and versatility in grasping task.

In future work, we will improve our QwenGrasp, aiming to conduct it in stacked scenes. Meanwhile to expand the function and make it to find the hidden object. Through means of improving the ability of the interaction between humans and robots, guide the robot arm to better conduct the grasping task by communicating. Furthermore, we will also try to use large model to guide the work of multiple robot arms, exploring the possibility of multi-robot collaboration.

%\clearpage
\bibliographystyle{IEEEtran}
\bibliography{QwenGrasp}

\end{document}